\documentclass{article}

\usepackage{microtype}
\usepackage{graphicx}
\usepackage{subcaption}
\usepackage{booktabs} 

\usepackage{hyperref}

\usepackage[accepted]{icml2026}

\usepackage{amsmath}
\usepackage{amssymb}
\usepackage{mathtools}
\usepackage{amsthm}
\usepackage{array}
\usepackage{xcolor}
\usepackage{colortbl}
\usepackage{listings}
\usepackage[capitalize,noabbrev]{cleveref}
\usepackage{multirow}
\usepackage{tabularx}

\theoremstyle{plain}

\theoremstyle{definition}

\theoremstyle{remark}

\lstdefinestyle{python}{
  language=Python,
  basicstyle=\ttfamily\scriptsize,
  keywordstyle=\color{blue},
  commentstyle=\color{gray},
  stringstyle=\color{red},
  numbers=left,
  numberstyle=\tiny,
  stepnumber=1,
  numbersep=6pt,
  breaklines=true,
  breakatwhitespace=false,
  frame=single,
  tabsize=2,
  showstringspaces=false
}
\usepackage[textsize=tiny]{todonotes}

\icmltitlerunning{LearniBridge: Learnable Calibration of Feature Caching for Diffusion Models Acceleration}

\begin{document}

\twocolumn[
  \icmltitle{
    LearniBridge: Learnable Calibration of Feature Caching\\
    for Diffusion Models Acceleration
  }



  \icmlsetsymbol{equal}{*}

  \begin{icmlauthorlist}
    \icmlauthor{Xuyue Huang}{thu}
    \icmlauthor{Zhe Chen}{comp}
    \icmlauthor{Wang Shen}{comp}
    \icmlauthor{Xiao-Ping Zhang}{thu}
  \end{icmlauthorlist}

  \icmlaffiliation{thu}{Shenzhen Ubiquitous Data Enabling Key Laboratory, Shenzhen International Graduate School, Tsinghua University, Shenzhen, China}
  \icmlaffiliation{comp}{Central Media Technology Institute, Huawei Technologies Co., Ltd., Shenzhen, China}
  \icmlcorrespondingauthor{Xiao-Ping Zhang}{xpzhang@ieee.org}

  \icmlkeywords{Machine Learning, ICML}

  \vskip 0.3in
]



\printAffiliationsAndNotice{}  

\begin{abstract}
Diffusion Transformers (DiTs) have driven substantial progress in image and video generation but suffer from prohibitive computational costs.
Feature caching accelerates inference by reusing intermediate representations.
Existing methods rely on historical features for implementation simplicity, 
yet suffer from severe error accumulation at high acceleration ratios.
To address this limitation, we investigate the nature of the requisite feature correction.
We demonstrate that the optimal calibration update is characterized by a shared low-rank subspace across diverse prompts.
Guided by this structural insight, we propose \textit{LearniBridge}, a learnable calibration mechanism for feature caching that bridges multiple timesteps through lightweight LoRA updates.
This mechanism enables effective calibration requiring only 3--5 training samples.
Extensive experiments on image and video generation show that \textit{LearniBridge} achieves up to $5.87\times$, $5.75\times$, and $4.10\times$ acceleration
on FLUX, HunyuanVideo, and WAN~2.1, respectively. On WAN~2.1, it improves VBench by 1.28\% over the previous SOTA at $4.10\times$ acceleration.
Our code is available at \url{https://github.com/Iiiiiiirene/LearniBridge}.

\end{abstract}

\section{Introduction}

Diffusion models~\cite{ho2020DM,dhariwal2021DM} have rapidly advanced generative modeling, achieving
state-of-the-art performance in image and video generation~\cite{rombach2022imageDM,blattmann2023videoDM}.
Recent work introduces Diffusion Transformers (DiTs)~\cite{peebles2023scalable,esser2024scaling,chen2023pixart} to further enhance generation quality.
However, these improvements come at the expense of high computational demands, which limit the practical applicability.

Various acceleration strategies~\cite{lu2022dpm,li2024svdquant} have been proposed to address these challenges.
Among them, feature caching~\cite{liu2025teacache,zhao2024PAB,yuan2024ditfastattn,zou2024toca,zhou2025easycache} effectively 
reduces inference costs by reusing intermediate representations from earlier timesteps.
Methods such as DeepCache\cite{ma2024deepcache} and FORA~\cite{selvaraju2024fora} exploit the similarity between adjacent timesteps for direct reuse.
To handle longer caching intervals, TaylorSeer~\cite{liu2025taylorseer} employs Taylor-series expansions over multi-step histories to predict future feature features.
However, TaylorSeer requires more cached features and relies on the assumption of smooth, higher-order differentiable feature trajectories that may not always hold.

Existing methods rely on historical features for implementation simplicity, yet incur severe error accumulation at high acceleration ratios.
This raises a critical question: Is there an underlying structure within these caching errors that enables efficient, learnable correction?
In this paper, by analyzing the spectral properties of feature correction matrices, we observe that the requisite updates consistently reside in a low-dimensional subspace.
This theoretically justifies that the calibration weights are naturally constrained by a low-rank structure.
Furthermore, we find that these low-rank subspaces remain remarkably consistent across diverse prompts, demonstrating a robust prompt-invariant structure.

Motivated by these observations, we introduce \textit{LearniBridge}, a learnable calibration mechanism for cached features, implemented via lightweight LoRA updates that bridge multiple timesteps.
The method consists of two phases: training and inference.
In the training phase, we optimize LoRA-based calibration weights to align cached features with their full-computation representations.
In the inference phase, full computation at the target timestep is bypassed, requiring only the inference of the LoRA-augmented final Transformer block. 
Due to the inherent low-rank nature of these calibration weights, our approach is highly parameter-efficient.
Furthermore, the prompt-invariant structure enables \textit{LearniBridge} to generalize broadly after training on minimal samples.
This design enables \textit{LearniBridge} to bridge multiple timesteps with minimal computational overhead.

Our main contributions are summarized as follows:
\begin{itemize}
  \item \textbf{Prompt-Invariant Low-Rank Structure:}
  We demonstrate that the optimal calibration updates in DiT linear layers reside in a prompt-invariant, low-dimensional subspace. 
  This finding provides a principled foundation for effective calibration.

  \item \textbf{LearniBridge:}
  We propose a LoRA-based framework that calibrates cached
  features to reconstruct future-step representations. 
  By using only 3-5 training prompts and lightweight low-rank updates, 
  it enables efficient integration with negligible computational overhead.

  \item \textbf{Outstanding performance:}
  \textit{LearniBridge} achieves up to $5.87\times$, $5.75\times$, and
  $4.10\times$ acceleration on FLUX, HunyuanVideo, and Wan~2.1, respectively,
  while maintaining high generation quality across image and video generation
  tasks.
\end{itemize}

\section{Related Work}
Diffusion models~\cite{ho2020DM,sohl2015deep} have achieved remarkable success in generative tasks. 
Early works predominantly employed U-Net structures~\cite{ronneberger2015u}, but struggled with scalability in larger models. 
This changed with the Diffusion Transformers (DiTs)~\cite{peebles2023scalable}, 
which resolved these constraints and enabled state-of-the-art performance in multiple domains ~\cite{chen2024pixart,chen2023pixart,yang2024cogvideox,zheng2024open}. 
However, sequential sampling leads to persistently high inference costs.
To address this, acceleration strategies have been extensively studied and are generally divided into Sampling Timestep Reduction and Denoising Network Acceleration.

\subsection{Diffusion Model Acceleration}

Extensive research focuses on accelerating diffusion models by either minimizing sampling steps or enhancing per-step efficiency. 
DDIM~\cite{song2020ddim} introduced a deterministic sampling paradigm that preserves generation fidelity with fewer iterations, 
a framework subsequently refined by the DPM-Solver series~\cite{lu2022dpm} through higher-order ODE solvers.
Consistency Models~\cite{song2023consistency} further establish self-consistent noise-to-data mappings, enabling one- or few-step generation. 
Additionally, model distillation~\cite{salimans2022progressive,meng2023distillation} 
compresses multi-step samplers into efficient student models that require significantly fewer denoising iterations.
To further alleviate the per-step computational burden, model quantization~\cite{kim2025ditto,li2023q,shang2023post} 
and structural pruning~\cite{fang2023structuralpruningdiffusionmodels,zhu2024dip} compress the diffusion backbones.
Beyond these approaches, feature caching represents a highly efficient acceleration paradigm due to its minimal training overhead and model-agnostic nature.

\subsection{Feature Caching}
Caching-based methods exploit the strong temporal coherence of intermediate
activations to skip redundant computation across diffusion timesteps.
DeepCache~\cite{song2023consistency} initially developed for U-Net architectures, reuses features across multiple steps.
Extending this concept to Transformer-based architectures, 
FORA~\cite{selvaraju2024fora} implements fundamental module-level output caching specifically for Diffusion Transformers (DiTs).
DiTFastAttn~\cite{yuan2024ditfastattn} further reduces costs 
by sharing attention outputs across spatial dimensions, time, and conditional branches. 
To maintain synthesis quality, ToCa~\cite{zou2024toca} incorporates dynamic feature updates to mitigate information loss caused by feature aging. 
In terms of cache decision mechanisms, 
TeaCache~\cite{liu2025teacache} introduces a calibrated polynomial estimator to predict output changes from input differences.
TaylorSeer~\cite{liu2025taylorseer} advances the paradigm from simple ``replication" to ``prediction" via 
a Taylor-series-inspired extrapolation scheme, significantly enhancing generation quality during long-range skipping.
Despite these advancements, 
existing methods predominantly rely on historical features for implementation simplicity,
which inevitably incurs severe error accumulation at high acceleration ratios.

\begin{figure*}[t]
  \centering
  \begin{minipage}[c]{0.80\textwidth}
    \centering
    \includegraphics[width=\linewidth]{SVD_global_100prompts_v4_step8.png}
    \captionof{figure}{SVD analysis of the aggregated input matrix $X_{t}^l$ across 100 distinct prompts.
    The singular values exhibit a rapid decay, indicating that spectral energy is concentrated in a few principal components.
    This implies that $X_{t}^l$ possesses an intrinsic low-rank structure, constraining the optimal correction $\Delta W^l$ to be low-rank.}
    \label{fig:SVD_all_layers}
  \end{minipage}
  \hfill
  \begin{minipage}[c]{0.19\textwidth}
    \centering
    \includegraphics[width=\linewidth]{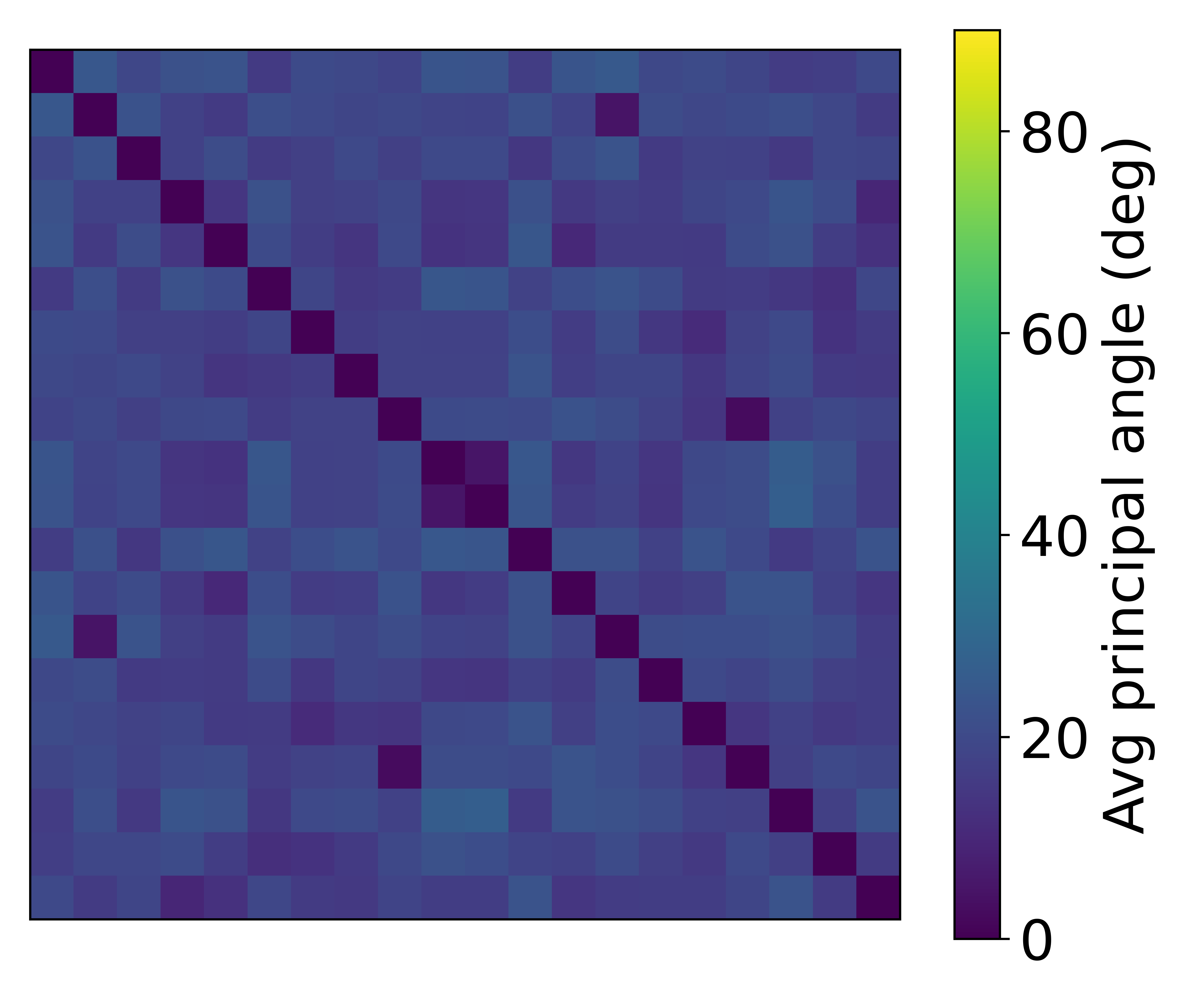}
    \captionof{figure}{Small angles between updates from disjoint prompt groups verify that the correction pattern is prompt-invariant.}    
    \label{fig:prompt_shared}
  \end{minipage}
\end{figure*}

\section{Method}
\subsection{Preliminary}

\paragraph{Diffusion Models.}
Diffusion models formulate generative modeling as learning to invert a gradual
noising process.
Starting from clean data $x_0 \sim q(x)$, a forward diffusion process constructs
a sequence $\{x_t\}_{t=1}^T$ by progressively adding Gaussian noise:
\begin{equation}
x_t = \sqrt{\alpha_t}\, x_{t-1} + \sqrt{1-\alpha_t}\, z_t,
\quad z_t \sim \mathcal{N}(0, I),
\end{equation}
where $\alpha_t \in (0,1]$ controls the signal-to-noise ratio at each timestep.
For appropriately chosen $\{\alpha_t\}_{t=1}^T$, the marginal distribution of
$x_T$ approaches an isotropic Gaussian.

The generative model parameterizes the reverse process using a neural network
that defines the conditional distributions:
\begin{equation}
p_\theta(x_{t-1} \mid x_t)
= \mathcal{N}\!\left(
x_{t-1}; \mu_\theta(x_t, t), \Sigma_\theta(x_t, t)
\right),
\end{equation}
and samples are obtained by iteratively applying these reverse transitions from
$t = T$ down to $t = 1$.
Since this procedure requires evaluating the backbone network at every
timestep, diffusion models typically incur substantial computational cost
during generation.

\paragraph{Diffusion Transformer Architecture.}
The Diffusion Transformers (DiTs) adopts a hierarchical architecture
$G = g_1 \circ \cdots \circ g_L$.
Each block is defined as
$g_l = F_{\mathrm{SA}}^l \circ F_{\mathrm{CA}}^l \circ F_{\mathrm{MLP}}^l$,
indicating that the input is sequentially processed by a feedforward (MLP) module,
a cross-attention module, and a self-attention module.
Each module is implemented in residual form as
$F_\alpha^l(x) = x + \operatorname{AdaLN}_\alpha^l\!\left(f_\alpha^l(x)\right)$
for $\alpha \in \{\mathrm{SA}, \mathrm{CA}, \mathrm{MLP}\}$.

The self-attention module computes the query, key, and value projections
$Q = x W_Q^l$, $K = x W_K^l$, and $V = x W_V^l$, followed by scaled dot-product
attention $\operatorname{Attn}(x) =
\operatorname{softmax}\!\left( Q K^{\top} / \sqrt{d_h} \right) V$ and an output
projection $f_{\mathrm{SA}}^l(x) = \operatorname{Attn}(x) W_O^l$.
The cross-attention module follows an identical formulation with its own
projection matrices. The MLP module consists of two linear layers with a
nonlinear activation, expressed as
$f_{\mathrm{MLP}}^l(x) = \sigma\!\left( x W_1^l \right) W_2^l$.
Together, these components define the standard DiT backbone that is applied
consistently across diffusion timesteps.

\paragraph{Feature Caching.}
Feature caching reduces computational cost by approximating block outputs across
diffusion timesteps. For block $l$ at timestep $t$, let $F^l\!\left(x_t^l\right)$
denote its output, and let $c_s^l = F^l\!\left(x_s^l\right)$ represent the cached
output at a reference timestep $s$. Given a set of reference timesteps
$S_{t,k} \subseteq \{t, \ldots, t-m\}$ with $k \in \{1, \ldots, m\}$, a general
caching scheme approximates the block output at timestep $t-k$ as:
\begin{equation}
\hat{F}^l\!\left(x_{t-k}^l\right)
= \Phi^l\!\left(\left\{c_s^l\right\}_{s \in S_{t,k}}, k\right),
\end{equation}
where $\Phi^l$ specifies the rule by which cached features are combined to
construct the approximation. Different caching methods instantiate $\Phi^l$
using different strategies, such as direct feature reuse or higher-order extrapolation.

\paragraph{Low-Rank Adaptation.}
Low-Rank Adaptation (LoRA) introduces trainable low-rank matrices into linear
layers to enable parameter-efficient fine-tuning~\cite{hu2022lora}. For a linear transformation
with weight $W^l \in \mathbb{R}^{d_{\text{in}} \times d_{\text{out}}}$, LoRA
augments the weight with a low-rank update:
\begin{equation}
\Delta W^l = B^l A^l ,
\end{equation}
where $A^l \in \mathbb{R}^{r \times d_{\text{out}}}$ and
$B^l \in \mathbb{R}^{d_{\text{in}} \times r}$, with
$r \ll \min(d_{\text{in}}, d_{\text{out}})$. During adaptation, the effective
weight becomes $W^l + \Delta W^l$, while the base weight $W^l$ remains frozen.
Only the low-rank factors $A^l$ and $B^l$ are updated, which substantially
reduces the number of trainable parameters and allows the adapted weights to be
stored and applied in a modular manner.

\begin{figure*}[t]
  \centering
  \includegraphics[width=0.98\textwidth]{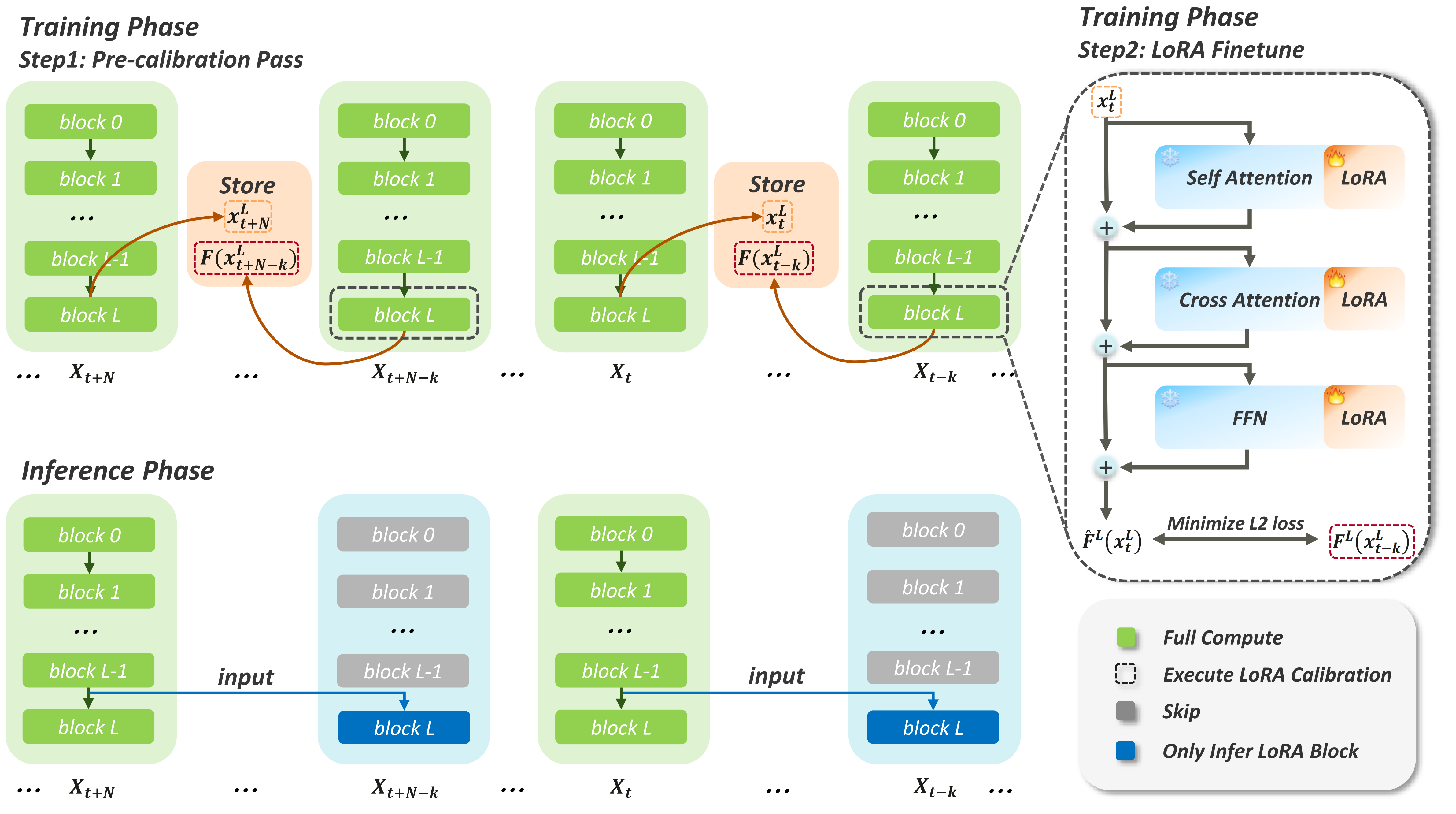}
    \caption{Overview of \textit{LearniBridge}. Our method consists of a training phase and an inference phase.
    During the training phase, a pre-calibration pass performs full computation at all timesteps, recording the final-block input $x_t^L$ and the corresponding ground-truth outputs $F^L(x_{t-k}^L)$ for calibrated timesteps.
    In the LoRA Finetune process, LoRA adapters are trained in the final block to map the cached input $x_t^L$ to the corresponding full-computation output $F^L(x_{t-k}^L)$.
    During the inference phase, full computation at the target timesteps is skipped, only infer the LoRA-augmented final Transformer block.}
    \label{fig:learnibridge_overview}
\end{figure*}

\subsection{Prompt-Invariant Low-Rank Calibration Update}
\label{sec:calibration_update}
Let $f^l(x_{t-k}^l)$ denote the output of a specific linear layer at the reference timestep $t-k$, 
where $x_{t-k}^l$ is the corresponding input.
We investigate the feature shift between the layer outputs given the previous input $x_t^l$ versus the current input $x_{t-k}^l$.
For each training sample $i$, we define the cross-timestep residual as:
\begin{equation}
e_{t \rightarrow t-k, i}^l \triangleq f^l\!\left(x_{t-k, i}^l\right) - f^l\!\left(x_{t, i}^l\right).
\end{equation}
We aggregate these residuals and the corresponding inputs from all $N$ samples into matrices $E_{t \rightarrow t-k}^l$ and $X_{t}^l$, respectively:
\begin{equation}
E_{t \rightarrow t-k}^l = \left[ e_{t \rightarrow t-k, 1}^l,\, \ldots,\, e_{t \rightarrow t-k, i}^l, \ldots, e_{t \rightarrow t-k, N}^l \right],
\end{equation}
\begin{equation}
X_{t}^l = \left[ x_{t,1}^l,\, \ldots, x_{t,i}^l,\, \ldots, \, x_{t,N}^l \right].
\end{equation}
Let $\Delta W^l$ represent the adaptive linear correction introduced to compensate for this discrepancy.
We model the relationship as $E_{t \rightarrow t-k}^l \approx \Delta W^l X_{t}^l$.
Minimizing the reconstruction error yields the closed-form solution:
\begin{equation}
\Delta W^l = E_{t \rightarrow t-k}^l (X_{t}^l)^\dagger,
\end{equation}
where $(X_{t}^l)^\dagger$ denotes the Moore-Penrose pseudo-inverse.

We next analyze the spectral properties of the input matrix $X_{t}^l$ via Singular Value Decomposition (SVD):
\begin{equation}
X_{t}^l = U \Sigma V^\top,
\end{equation}
where $U$ and $V$ are orthogonal matrices, and $\Sigma$ is a diagonal matrix of singular values.
Substituting the pseudo-inverse expression $(X_{t}^l)^\dagger = V \Sigma^+ U^\top$ into the solution for $\Delta W^l$, we obtain:
\begin{equation}
\Delta W^l = E_{t \rightarrow t-k}^l V \Sigma^+ U^\top.
\end{equation}
As illustrated in Figure~\ref{fig:SVD_all_layers}, we constructed $X_{t}^l$ using samples from 100 distinct prompts.
The singular values of $X_{t}^l$ exhibit a rapid decay, 
indicating that the spectral energy is concentrated in a few principal components.
This empirical observation suggests that $X_{t}^l$ possesses an intrinsic low-rank structure.
Given that the rank of a matrix product is constrained by the minimum rank of its factors, it follows that:
\begin{equation}
\text{rank}(\Delta W^l) \le \text{rank}((X_{t}^l)^\dagger) = \text{rank}(X_{t}^l) \le r.
\end{equation}
This implies that the optimal correction $\Delta W^l$ is inherently low-rank.

To verify the universality of the update direction, we randomly partition 100 prompts into 20 disjoint groups 
and compute the optimal $\Delta W^l$ for each group.
For each $\Delta W^l$, we perform SVD to obtain its principal subspace, and Figure~\ref{fig:prompt_shared} visualizes the pairwise angles between these subspaces across different prompt groups.
The consistently small angles reveal a strong structural similarity between these updates, implying that the required feature correction follows a universal pattern,
independent of the specific input prompts.

\begin{table*}[t]
\centering
\caption{Quantitative comparison in text-to-image generation on DrawBench with FLUX.1-dev.}
\label{tab:flux}
\fontsize{8pt}{9pt}\selectfont
\setlength{\tabcolsep}{3.5pt}
\renewcommand{\arraystretch}{1.08}
\begin{tabular}{l|cccc|ccccc}
\toprule
\multicolumn{1}{l|}{\textbf{Method}} &
\multicolumn{4}{c|}{\textbf{Acceleration}} &
\multicolumn{5}{c}{\textbf{Quality Metrics}} \\
\textbf{FLUX.1-dev} &
\textbf{Latency(s)$\downarrow$} & \textbf{Speed$\uparrow$} &
\textbf{FLOPs(T)$\downarrow$} & \textbf{Speed$\uparrow$} &
\textbf{ImageReward$\uparrow$} & \textbf{CLIP$\uparrow$} &
\textbf{PSNR$\uparrow$} & \textbf{SSIM$\uparrow$} &
\textbf{LPIPS$\downarrow$} \\
\midrule
Original
& 27.32 & $1.00\times$ & 3719.5 & $1.00\times$
& 0.9885 & 0.8102 & $-$ & $-$ & $-$ \\
22\% steps
& 6.00 & $4.55\times$ & 817.5 & $4.55\times$
& 0.8669 & 0.8130 & 25.9587 & 0.6720 & 0.3691 \\
\midrule
ToCA ($N{=}9$)~\cite{zou2024toca}
& 6.88 & $3.97\times$ & 854.4 & $4.35\times$
& 0.8352 & 0.8045 & 27.9813 & 0.7012 & 0.3155 \\
TeaCache ($\delta{=}0.8$)~\cite{liu2025teacache}
& 6.60 & $4.14\times$ & 892.0 & $4.17\times$
& 0.8975 & 0.8103 & 28.6508 & 0.7350 & \textbf{0.2538} \\
TaylorSeer ($N{=}5,O{=}2$)~\cite{liu2025taylorseer}
& 6.20 & $4.41\times$ & 817.5 & $4.55\times$
& 0.9359 & \textbf{0.8164} & 29.8558 & 0.7625 & 0.2697 \\
\rowcolor{gray!20}
\textbf{LearniBridge} ($N{=}5$)
& 6.27 & $4.36\times$ & 839.6 & $4.43\times$
& \textbf{0.9590} & 0.8128 & \textbf{30.1525} & \textbf{0.7879} & 0.2682 \\
\midrule
ToCA ($N{=}10$)~\cite{zou2024toca}
& 5.78 & $4.73\times$ & 714.7 & $5.20\times$
& 0.7998 & 0.7956 & 26.9854 & 0.6390 & 0.3702 \\
TeaCache ($\delta{=}1$)~\cite{liu2025teacache}
& 5.66 & $4.83\times$ & 743.6 & $4.89\times$
& 0.8398 & 0.8060 & 27.0821 & 0.6996 & 0.3702 \\
TaylorSeer ($N{=}6,O{=}2$)~\cite{liu2025taylorseer}
& 5.47 & $4.99\times$ & 745.4 & $4.99\times$
& 0.9033 & 0.8094 & 28.7006 & 0.7191 & 0.3109 \\
\rowcolor{gray!20}
\textbf{LearniBridge} ($N{=}6$)
& 5.61 & $4.87\times$ & 759.1 & $4.90\times$
& \textbf{0.9133} & \textbf{0.8364} & \textbf{29.7491} & \textbf{0.7407} & \textbf{0.3021} \\
\midrule
ToCA ($N{=}12$)~\cite{zou2024toca}
& 4.65 & $5.87\times$ & 628.3 & $5.92\times$
& 0.7019 & 0.7826 & 26.4802 & 0.5856 & 0.3928 \\
TeaCache ($\delta{=}1.4$)~\cite{liu2025teacache}
& 4.56 & $5.99\times$ & 603.8 & $6.16\times$
& 0.7252 & 0.8026 & 26.5802 & 0.6338 & 0.3928 \\
TaylorSeer ($N{=}8,O{=}2$)~\cite{liu2025taylorseer}
& 4.48 & $6.10\times$ & 596.1 & $6.24\times$
& 0.8212 & 0.8041 & 26.8228 & 0.6750 & 0.3647 \\
\rowcolor{gray!20}
\textbf{LearniBridge} ($N{=}8$)
& 4.43 & $6.17\times$ & 599.9 & $6.20\times$
& \textbf{0.8308} & \textbf{0.8164} & \textbf{28.3464} & \textbf{0.6870} & \textbf{0.3549} \\
\bottomrule
\end{tabular}
\end{table*}

\subsection{LoRA-Based Calibration Architectures}
Let $F^l(x_{t-k}^l)$ denote the output of a specific block at the reference timestep $t-k$.
Leveraging the low-rank update structure, the output at a later timestep
can be approximated from the input at an earlier timestep as:
\begin{equation}
F^l\!\left(x_{t-k}^l ; W^l\right)
\approx F^l\!\left(x_t^l ; W^l + \Delta W^l\right).
\end{equation}

This relation indicates that a suitably parameterized low-rank adapter can
reproduce the skipped-timestep representation using only cached features.

As illustrated in Figure~\ref{fig:learnibridge_overview}, \textit{LearniBridge} implements this insight by applying a lightweight residual correction to the final Transformer block $g_L$ 
(composed as $g_l = F_{\mathrm{SA}}^l \circ F_{\mathrm{CA}}^l \circ F_{\mathrm{MLP}}^l$). 
This mechanism compensates for the temporal feature shift by modeling it as a learnable update.
Restricting LoRA to $g_L$ minimizes trainable parameters and memory footprint while preserving the backbone architecture. 
Since no auxiliary blocks are executed, this design incurs negligible inference latency, yielding a plug-and-play calibration module.

For any linear transformation $W^l$, LoRA introduces a low-rank update:
\begin{equation}
\Delta W^l = B^l A^l, \qquad
r \ll \min\!\left(d_{\mathrm{in}}, d_{\mathrm{out}}\right),
\end{equation}
and replaces $W^l$ with $W^l + \Delta W^l$ while keeping the base weight $W^l$
itself frozen. \textit{LearniBridge} applies this parameterization exclusively to the
final block $g_L$, introducing low-rank updates
$\Delta W_Q^L, \Delta W_K^L, \Delta W_V^L, \Delta W_O^L, \Delta W_1^L,$ and
$\Delta W_2^L$ to all linear layers within the block.

\textbf{Training Phase}\\
During pre-calibration pass, we collect a small set of prompts (3--5) and run full diffusion trajectories.
For each fully computed timestep $t$, we record the input to the final block, denoted as $x_t^L$.
For timesteps $x_{t-k}^L$ that will be calibrated during inference, we record corresponding ground-truth outputs $F^L(x_{t-k}^L)$.

During LoRA finetune process, the LoRA parameters in $g_L$ are trained while keeping all base weights frozen.
The objective is to ensure that the LoRA-augmented final block, when taking the
cached feature $x_t^L$ as input, can well approximate the target output
$F^L\!\left(x_{t-k}^L\right)$ obtained under full computation. Over all training
pairs $(t,t-k)$, we minimize:
\begin{equation}
\mathcal{L}_{\text{LearniBridge}}
= \sum_i
\left\|
\hat{F}^L\!\left(x_{t,i}^L\right)
- F^L\!\left(x_{t-k,i}^L\right)
\right\|_2^2.
\end{equation}
Here, $\hat{F}^L\!\left(x_{t,i}^L\right)$ serves as the output of LoRA-augmented final block. 
By modeling the temporal shift, we reconstruct representations at skipped timesteps directly from the cached features.

\textbf{Inference Phase}\\
During inference, the model periodically executes full computation
at intervals of $N$ timesteps and caches the corresponding input to the final block,
denoted as $x_t^L$. For a target timestep $t-k$, we retrieve the cached feature $x_t^L$ 
from the nearest full-compute step $t$.
Instead of recomputing all blocks
$g_1, \ldots, g_{L-1}$ at timestep $t-k$, \textit{LearniBridge} directly feeds this cached
feature $x_t^L$ into the LoRA-augmented final block $g_L$, producing an
approximation $\hat{F}^L\!\left(x_t^L\right)$ of the full computation output.
This substitution allows the model to bypass earlier blocks, 
relying on the trained low-rank correction to recover cross-timestep feature evolution.

\begin{table*}[t]
\centering
\caption{Quantitative comparison in text-to-video generation for HunyuanVideo on VBench.}
\label{tab:hunyuan}
\fontsize{8pt}{9pt}\selectfont
\renewcommand{\arraystretch}{1.08}
\begin{tabular}{l|cccc|cccc}
\toprule
\multicolumn{1}{l|}{\textbf{Method}} &
\multicolumn{4}{c|}{\textbf{Acceleration}} &
\multicolumn{4}{c}{\textbf{Quality Metrics}} \\
\textbf{HunyuanVideo} &
\textbf{Latency(s)$\downarrow$} & \textbf{Speed$\uparrow$} &
\textbf{FLOPs(T)$\downarrow$} & \textbf{Speed$\uparrow$} &
\textbf{VBench(\%)} & \textbf{PSNR$\uparrow$} &
\textbf{SSIM$\uparrow$} & \textbf{LPIPS$\downarrow$} \\
\midrule
Original
& 617.79 & $1.00\times$ & 29773.0 & $1.00\times$
& 80.93 & $-$ & $-$ & $-$ \\
22\% steps
& 135.78 & $4.55\times$ & 6550.1 & $4.55\times$
& 78.88 & 15.09 & 0.5669 & 0.3997 \\
\midrule
TeaCache ($\delta{=}0.3$)~\cite{liu2025teacache}
& 167.88 & $3.68\times$ & 7794.0 & $3.82\times$
& 80.07 & 18.36 & 0.7155 & 0.2629 \\
TaylorSeer ($N{=}4,O{=}1$)~\cite{liu2025taylorseer}
& 161.30 & $3.83\times$ & 7733.2 & $3.85\times$
& 80.74 & 19.47 & 0.6686 & 0.3096 \\
\rowcolor{gray!20}
\textbf{LearniBridge} ($N{=}4$)
& 164.74 & $3.75\times$ & 7876.5 & $3.78\times$
& \textbf{80.84} & \textbf{20.55} & \textbf{0.7508} & \textbf{0.2314} \\
\midrule
ToCa ($N{=}5$)~\cite{zou2024toca}
& 149.94 & $4.12\times$ & 7005.4 & $4.25\times$
& 79.25 & 18.13 & 0.6002 & 0.3885 \\
TeaCache ($\delta{=}0.4$)~\cite{liu2025teacache}
& 130.89 & $4.72\times$ & 6151.5 & $4.84\times$
& 79.40 & 16.91 & 0.6649 & 0.3372 \\
TaylorSeer ($N{=}5,O{=}1$)~\cite{liu2025taylorseer}
& 132.86 & $4.65\times$ & 5966.5 & $4.99\times$
& \textbf{80.13} & 18.28 & 0.6121 & 0.3722 \\
\rowcolor{gray!20}
\textbf{LearniBridge} ($N{=}5$)
& 125.31 & $4.93\times$ & 5966.5 & $4.99\times$
& 79.93 & \textbf{18.80} & \textbf{0.7396} & \textbf{0.2349} \\
\midrule
TaylorSeer ($N{=}7,O{=}1$)~\cite{liu2025taylorseer}
& 103.30 & $5.98\times$ & 4771.3 & $6.24\times$
& 79.14 & 17.77 & 0.6122 & 0.4232 \\
\rowcolor{gray!20}
\textbf{LearniBridge} ($N{=}7$)
& 100.45 & $6.15\times$ & 4794.4 & $6.21\times$
& \textbf{79.51} & \textbf{17.84} & \textbf{0.6504} & \textbf{0.3780} \\
\bottomrule
\end{tabular} 
\end{table*}

\begin{figure}[t]
  \centering
  \includegraphics[width=\columnwidth]{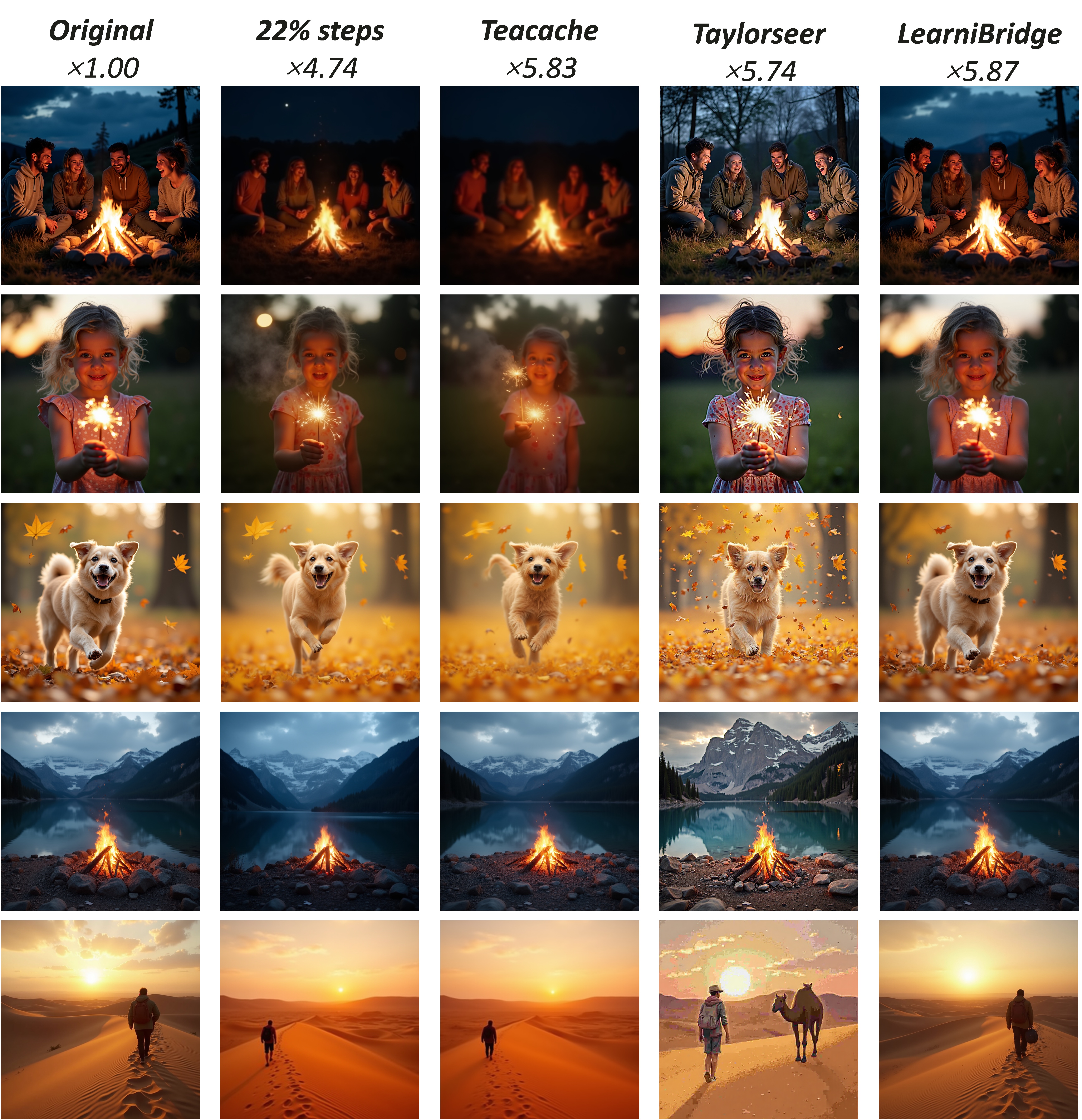}
  \caption{Detailed visualization results for different acceleration methods on FLUX.1-dev. Existing methods suffer from severe content deviation, blurring artifacts, or abnormal color contrast at high speedup, 
whereas \textit{LearniBridge} maintains high content fidelity and superior visual quality even at nearly $6\times$ acceleration.}
  \label{fig:flux_dev}
\end{figure}

\section{Experiments}
\subsection{Experiment Settings}
\paragraph{Model Configurations.}
Experiments are conducted on three state-of-the-art visual generative models:
the text-to-image generation model FLUX.1-dev~\cite{blackforestlabs2024flux}, and the text-to-video generation
models HunyuanVideo~\cite{kong2024hunyuanvideo} and WAN~2.1-1.3B~\cite{wan2025wan}.
All images and videos used for both quantitative and qualitative evaluation are generated on Ascend~910B devices.

\paragraph{Evaluation and Metrics.}
For text-to-image generation, we generate 200 images using prompts from the DrawBench benchmark~\cite{saharia2022photorealistic}.
Image quality and text--image alignment are evaluated using ImageReward~\cite{xu2023imagereward} and
CLIP Score~\cite{hessel2021clipscore}. For text-to-video generation, we produce a total of 4{,}730 videos by generating five samples for each of the 946 prompts. 
Model performance is comprehensively evaluated using the VBench~\cite{huang2024vbench} framework. 
Additionally, the fidelity of the generated outputs with respect to the original results is quantitatively assessed using PSNR, SSIM~\cite{wang2004ssim}, and LPIPS~\cite{zhang2018lpips}.

\subsection{Text-to-Image Generation}
\label{sec:e}
\paragraph{Quantitative Study.}

As shown in Table~\ref{tab:flux}, we compare \textit{LearniBridge} with existing acceleration methods,
including ToCa, TeaCache, and TaylorSeer, on the FLUX.1-dev model.
Under moderate acceleration, \textit{LearniBridge} achieves a $4.36\times$ speedup while
maintaining strong semantic and visual quality, with IR ($0.9590\uparrow$),
CLIP ($0.8128\uparrow$), PSNR ($30.1525\uparrow$), and SSIM ($0.7879\uparrow$).
It outperforms ToCa, TaylorSeer and remains competitive with TeaCache across
most metrics.
At a $4.98\times$ speedup, \textit{LearniBridge} achieves the highest CLIP score
($0.8364\uparrow$) while maintaining strong perceptual quality.
Even under aggressive acceleration at $5.87\times$, \textit{LearniBridge} preserves
highest IR ($0.8308\uparrow$), indicating superior robustness as the acceleration factor
increases.

\paragraph{Qualitative Study.}
Figure~\ref{fig:flux_dev} presents a visual comparison between \textit{LearniBridge} and baseline methods
on FLUX.1-dev.
When the speedup approaches $6\times$, TeaCache exhibits significant content
deviation from the original outputs, accompanied by noticeable blurring
artifacts.
TaylorSeer produces results with low similarity to the original images and
suffers from abnormal color contrast.
In contrast, \textit{LearniBridge} preserves high content consistency with the original
outputs while maintaining substantially better visual fidelity, demonstrating
its effectiveness under large timestep skipping.

\begin{figure*}[t]
  \centering
  \includegraphics[width=\textwidth]{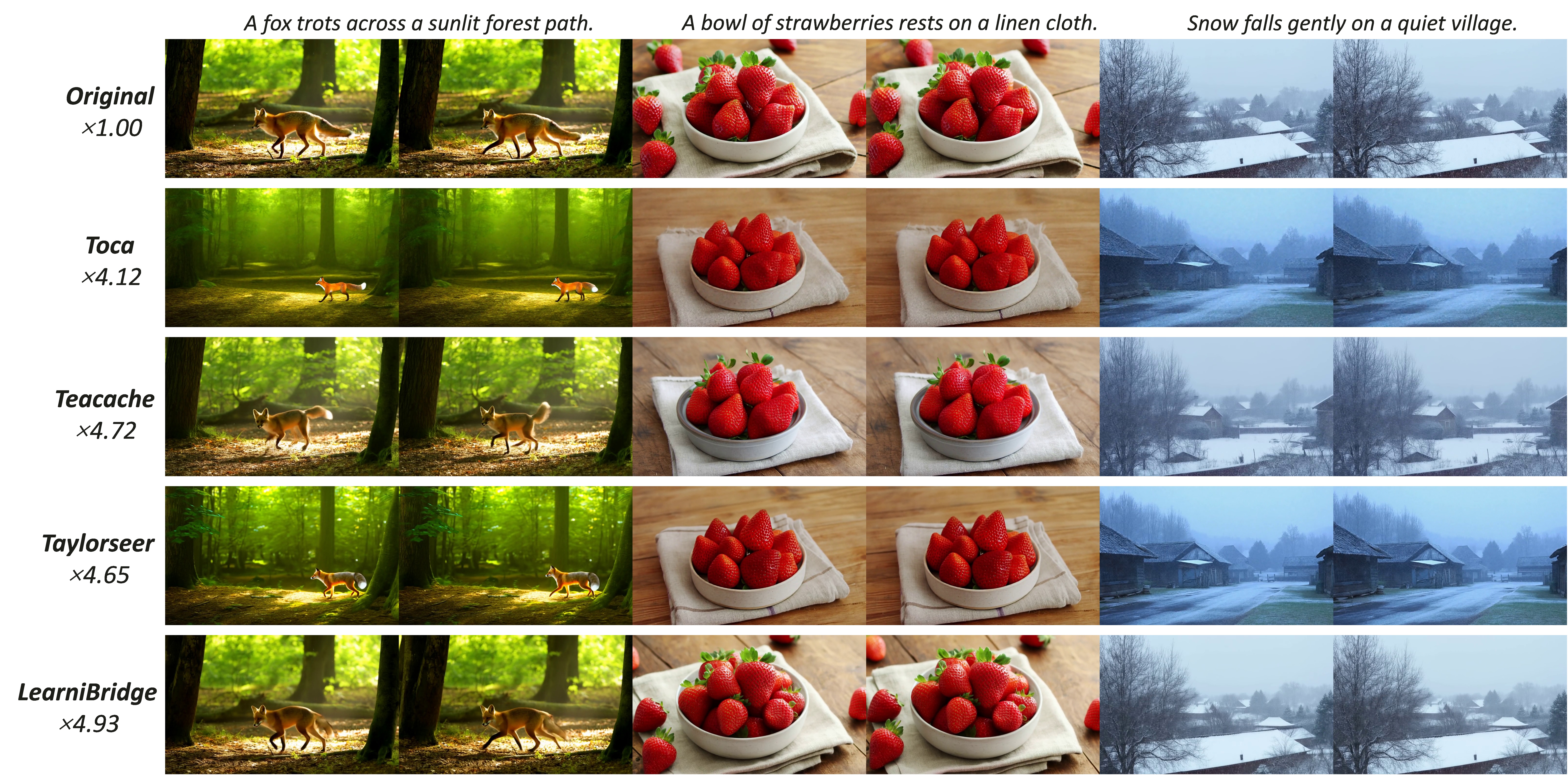}
  \caption{Visualization of different acceleration methods on HunyuanVideo. 
  While achieving higher acceleration ratios, other methods exhibit issues such as motion detail loss, content deviation, visual quality degrade. In contrast, our method maintains high-quality generation without these problems.}
  \label{fig:hunyuan}
\end{figure*}
\begin{figure}[t]
  \centering
  \includegraphics[width=\columnwidth]{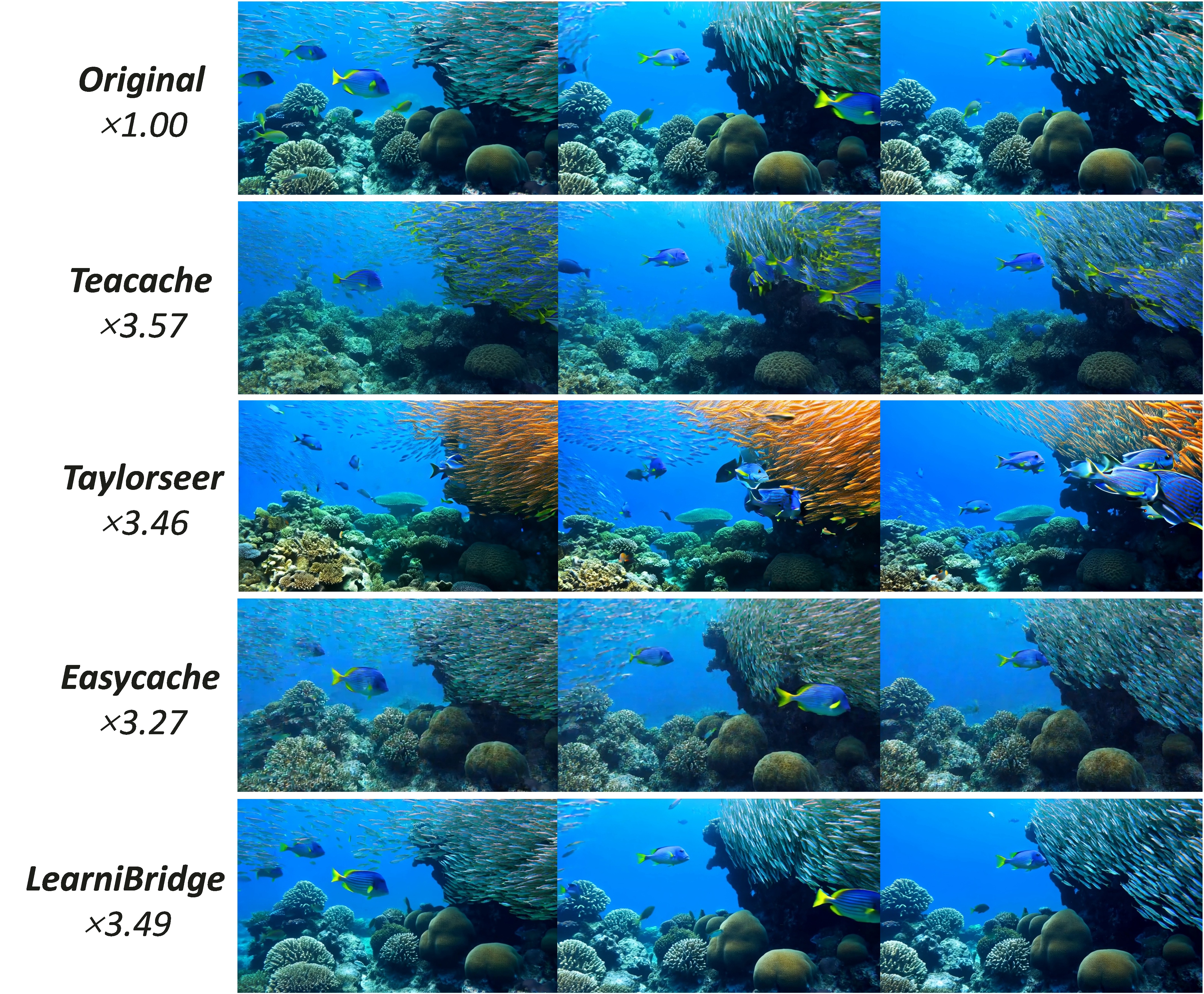}
  \caption{Visualization of different acceleration methods on WAN 2.1-1.3B.
  Baseline methods exhibit inconsistent color reproduction, degraded motion quality,
  and visible blurring, while \textit{LearniBridge} maintains high visual quality close to the original video.}
  \label{fig:wan}
\end{figure}

\subsection{Text-to-Video Generation}
\paragraph{Quantitative Study.}

As shown in Table~\ref{tab:hunyuan}, we compare \textit{LearniBridge} with existing acceleration methods,
including ToCa, TeaCache, and TaylorSeer, on the HunyuanVideo model.
Under moderate acceleration, \textit{LearniBridge} achieves a $3.75\times$ speedup with a
VBench score of $80.84$, outperforming TeaCache and TaylorSeer at comparable
speedup levels.
Even under more aggressive acceleration at $5.75\times$, \textit{LearniBridge}
consistently maintains higher VBench scores and better perceptual metrics than
TaylorSeer.

As shown in Table~\ref{tab:wan21}, we evaluate \textit{LearniBridge} on the WAN~2.1-1.3B model and
compare it with EasyCache, TeaCache, and TaylorSeer.
Existing approaches are generally limited to approximately $3\times$ speedup
and incur substantial degradation in output quality.
In contrast, \textit{LearniBridge} achieves speedup factors of $3.49\times$ (VBench
$81.21$) and $4.10\times$ (VBench $80.51$), while consistently outperforming all
competing methods across all quality metrics.
Notably, at a $3.49\times$ speedup, \textit{LearniBridge} preserves strong visual
fidelity, achieving PSNR ($21.26\uparrow$), SSIM ($0.8240\uparrow$), and LPIPS
($0.1824\downarrow$).

\begin{table*}[t] 
\centering
\caption{Quantitative comparison in text-to-video generation for WAN 2.1-1.3B on VBench.}
\label{tab:wan21}
\fontsize{8pt}{9pt}\selectfont
\renewcommand{\arraystretch}{1.08}
\begin{tabular}{l|cccc|cccc}
\toprule
\multicolumn{1}{l|}{\textbf{Method}} &
\multicolumn{4}{c|}{\textbf{Acceleration}} &
\multicolumn{4}{c}{\textbf{Quality Metrics}} \\
\textbf{WAN 2.1-1.3B} &
\textbf{Latency(s)$\downarrow$} & \textbf{Speed$\uparrow$} &
\textbf{FLOPs(T)$\downarrow$} & \textbf{Speed$\uparrow$} &
\textbf{VBench(\%)} & \textbf{PSNR$\uparrow$} &
\textbf{SSIM$\uparrow$} & \textbf{LPIPS$\downarrow$} \\
\midrule
Original
& 291.55 & $1.00\times$ & 13996.0 & $1.00\times$
& 81.52 & $-$ & $-$ & $-$ \\
\midrule
EasyCache ($\delta{=}0.13$)~\cite{zhou2025easycache}
& 89.16 & $3.27\times$ & 4203.0 & $3.33\times$
& 79.61 & 13.77 & 0.4745 & 0.4417 \\
TaylorSeer ($N{=}4,O{=}2$)~\cite{liu2025taylorseer}
& 84.26 & $3.46\times$ & 4760.5 & $2.94\times$
& 79.17 & 14.84 & 0.4456 & 0.4425 \\
TeaCache ($\delta{=}0.2$)~\cite{liu2025teacache}
& 81.67 & $3.57\times$ & 3898.6 & $3.59\times$
& 80.04 & 18.24 & 0.6804 & 0.3002 \\
\rowcolor{gray!20}
\textbf{LearniBridge} ($N{=}4$)
& 83.54 & $3.49\times$ & 3802.3 & $3.68\times$
& \textbf{81.21} & \textbf{21.26} & \textbf{0.8240} & \textbf{0.1824} \\
\midrule
TeaCache ($\delta{=}0.3$)~\cite{liu2025teacache}
& 73.25 & $3.98\times$ & 3364.4 & $4.16\times$
& 79.23 & 12.82 & 0.4691 & 0.4327 \\
\rowcolor{gray!20}
\textbf{LearniBridge} ($N{=}5$) 
& 71.11 & $4.10\times$ & 3180.9 & $4.40\times$
& \textbf{80.51} & \textbf{16.32} & \textbf{0.6254} & \textbf{0.3228} \\
\bottomrule
\end{tabular}
\end{table*}

\paragraph{Qualitative Study.}
Figure~\ref{fig:hunyuan} and Figure~\ref{fig:wan} present qualitative visual
comparisons between \textit{LearniBridge} and representative baseline methods on
HunyuanVideo and WAN~2.1-1.3B, respectively.
As illustrated in Figure~\ref{fig:hunyuan}, TaylorSeer and ToCa often introduce
severe content inconsistency across frames. For example, the appearance of the
fox, the shape of the strawberries, and the houses in snowy scenes undergo
significant and unrealistic changes over time.
TeaCache mainly suffers from degraded temporal dynamics, leading to unsmooth and
unnatural motion. In Figure~\ref{fig:hunyuan}, the fox exhibits abnormal motion
artifacts, such as the emergence of an extra leg. Similarly, in
Figure~\ref{fig:wan}, the school-of-fish scene becomes fragmented, resulting in
noticeably discontinuous motion.
EasyCache shows evident visual instability when the acceleration ratio exceeds
$3\times$, manifesting as frame-wise flickering and floating artifacts that
significantly deteriorate visual quality.
In contrast, \textit{LearniBridge} consistently preserves high fidelity to the original
video content while maintaining strong temporal coherence, producing smoother,
more stable, and visually consistent dynamic video generation results.

\begin{figure}[t]
  \centering
  \begin{subfigure}[t]{0.49\columnwidth}
    \centering
    \includegraphics[width=\linewidth]{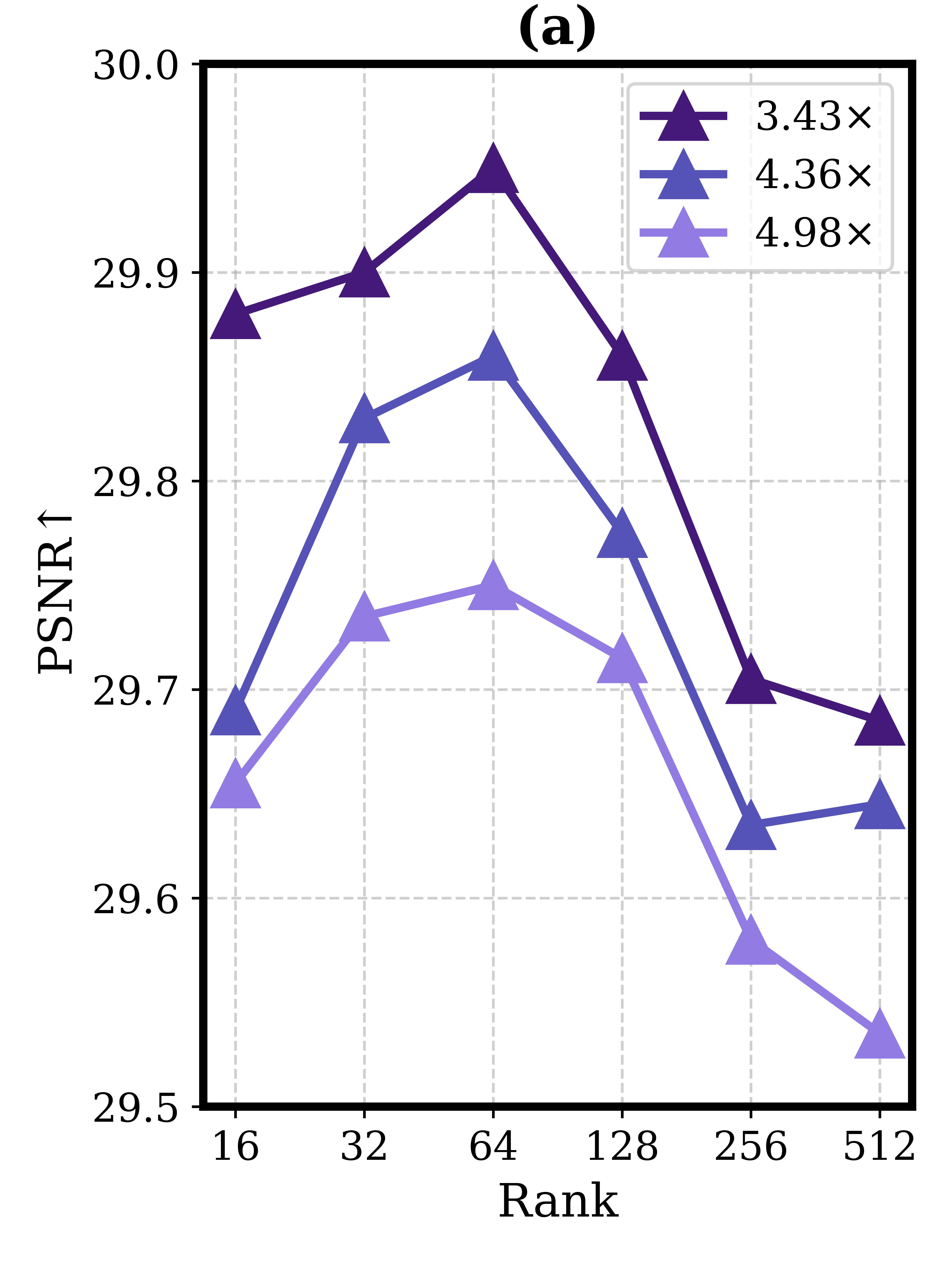}
    \label{fig:abl_rank}
  \end{subfigure}
  \hfill
  \begin{subfigure}[t]{0.49\columnwidth}
    \centering
    \includegraphics[width=\linewidth]{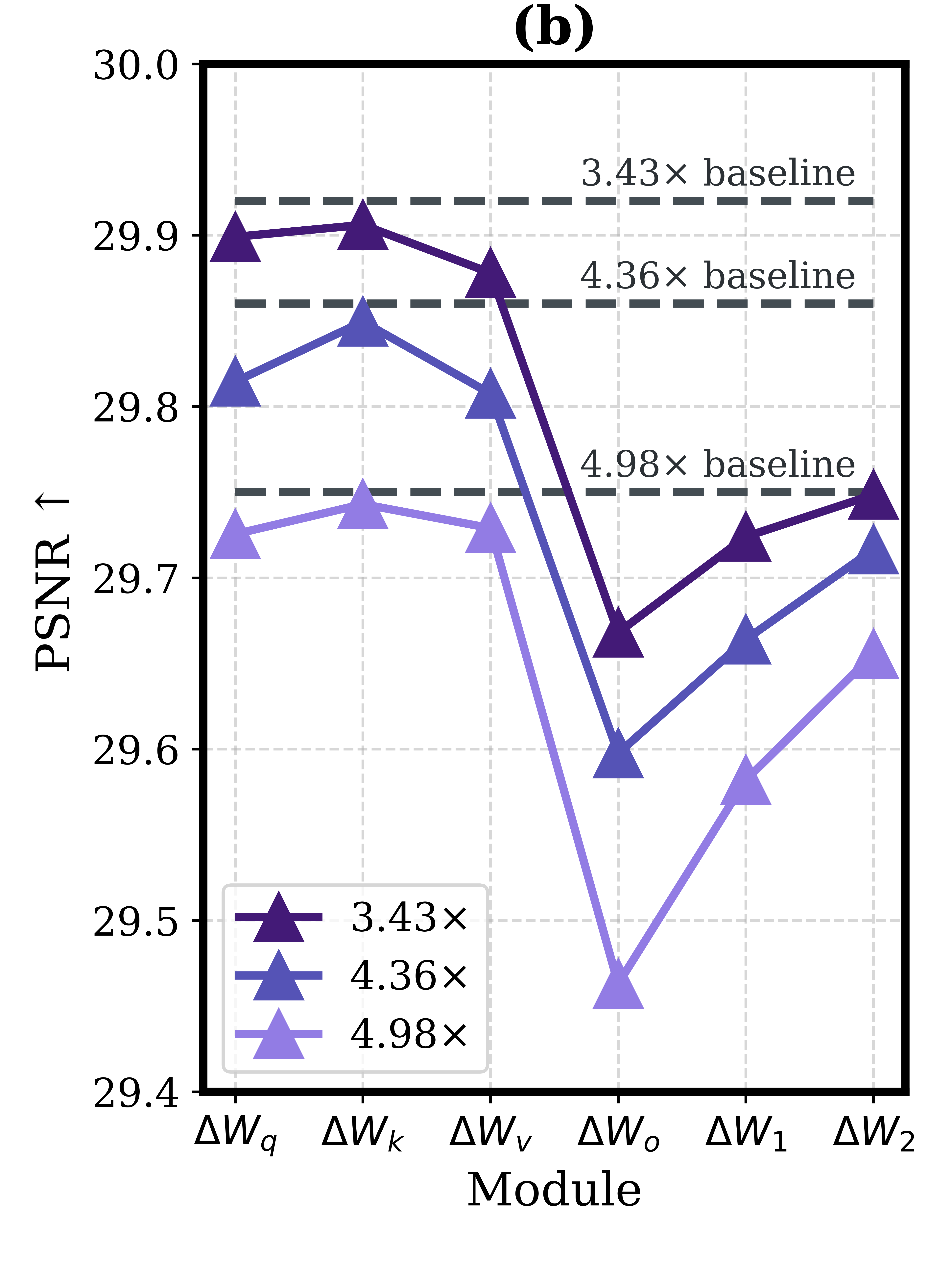}
    \label{fig:abl_module}
  \end{subfigure}
  \caption{
  \textbf{(a)} Impact of varying the rank of LoRA adapters.
  As the rank increases, reconstruction quality first improves and then degrades,
  indicating that larger ranks do not necessarily lead to better calibration.
  \textbf{(b)} Impact of selectively removing LoRA adapters from different linear layers.
  All modules contribute to preserving high-fidelity reconstruction after acceleration.}
  \label{fig:abl_all}
\end{figure}

\section{Ablation Studies}
\paragraph{Impact of Varying Rank}

In this section, we evaluate the impact of varying the rank of the LoRA adapters on acceleration performance, 
with detailed results presented in Figure~\ref{fig:abl_all}(a).
We observe that the reconstruction quality, measured by PSNR, improves steadily as the rank increases, reaching a peak at $r=64$.
Beyond this point, further increasing the rank yields diminishing returns, 
indicating that a rank of $64$ provides sufficient capacity to model the feature discrepancy between timesteps.
This observation confirms that the calibration task can be effectively solved with a moderate parameter budget.

\paragraph{Impact of Varying Module}

In the experiments presented in Section~4, LoRA adapters are applied to all
linear layers, including the query, key, value, and output projections, as well
as the feed-forward layers, corresponding to $\Delta W_q$, $\Delta W_k$, $\Delta W_v$, $\Delta W_o$, $\Delta W_1$,
and $\Delta W_2$.
We investigate the impact of selectively removing individual LoRA adapters under
different acceleration ratios to analyze how excluding a specific linear layer
affects overall performance, as illustraed in Figure~\ref{fig:abl_all}(b).
Reconstruction quality is quantitatively evaluated using PSNR.
Experimental results show that removing the LoRA adapters associated with the
query, key, and value projection layers leads to a slight degradation in
similarity to the original images.
In contrast, excluding the adapters corresponding to the output projection
($W_o$) and the feed-forward network layers ($W_1$ and $W_2$) results in a more
pronounced performance drop.
This indicates that the dense feature transformations are more critical for calibration than the attention routing components.

\begin{table}[t]
\centering
\caption{Ablation on calibration prompt length.}
\label{tab:ablation_prompt_length}
\footnotesize
\begin{tabular}{lccc}
\toprule
\textbf{Prompt Length} & \textbf{PSNR$\uparrow$} & \textbf{SSIM$\uparrow$} & \textbf{LPIPS$\downarrow$} \\
\midrule
Short (0--5 words) & 29.6247 & 0.6927 & 0.3135 \\
Medium (10--15 words) & \textbf{30.1569} & \textbf{0.7898} & 0.2654 \\
Long (20--25 words) & 30.0137 & 0.7878 & \textbf{0.2594} \\
\bottomrule
\end{tabular}
\end{table}

\begin{table}[t]
\centering
\caption{Ablation on the number of calibration prompts.}
\label{tab:ablation_prompt_number}
\footnotesize
\begin{tabular}{lccc}
\toprule
\textbf{Number of Prompts} & \textbf{PSNR$\uparrow$} & \textbf{SSIM$\uparrow$} & \textbf{LPIPS$\downarrow$} \\
\midrule
1 & 27.7913 & 0.6771 & 0.3829 \\
3 & 29.1079 & 0.6970 & 0.3184 \\
5 & 30.1623 & \textbf{0.7799} & 0.2681 \\
10 & 30.0728 & 0.7765 & \textbf{0.2635} \\
20 & \textbf{30.2297} & 0.7679 & 0.2746 \\
30 & 30.0814 & 0.7642 & 0.2699 \\
\bottomrule
\end{tabular}
\end{table}

\paragraph{Impact of Calibration Prompts}
We further study the influence of calibration prompts from two aspects: prompt length and prompt number. 
As shown in Table~\ref{tab:ablation_prompt_length}, medium and long prompts consistently outperform short prompts, with short prompts yielding notably lower SSIM and higher LPIPS. This indicates that overly simple inputs provide insufficient information for learning an effective calibration.
Table~\ref{tab:ablation_prompt_number} reports the effect of varying the number of calibration prompts. Performance improves substantially when increasing the number of prompts from 1 to 5, suggesting that a small prompt set is sufficient for calibration. Beyond 5 prompts, the performance largely saturates and only exhibits minor fluctuations.

\section{Conclusion}
In this paper, we present \textit{LearniBridge}, a LoRA-based calibration method for cached features that accelerates diffusion models.
We addressed the computational inefficiencies of Diffusion Transformers (DiTs) 
by investigating the nature of feature correction in acceleration tasks. 
Our analysis revealed that the requisite updates for feature reuse are characterized 
by a shared low-rank subspace across diverse prompts.
Building on this, we proposed \textit{LearniBridge}, a LoRA-based method that 
effectively calibrates historical features to bridge timestep discrepancies. 
Extensive experiments demonstrate that \textit{LearniBridge} achieves up to $5.87\times$, 
$5.75\times$, and $4.10\times$ acceleration on FLUX, HunyuanVideo, and WAN~2.1, respectively, 
while preserving high-quality generation capabilities.

\section*{Acknowledgment}
This work is supported by the National Natural Science Foundation of
China under Grant 62388102, by the Shenzhen Ubiquitous Data Enabling
Key Lab under Grant ZDSYS20220527171406015, and by the Tsinghua Shenzhen
International Graduate School-Shenzhen Pengrui Endowed Professorship
Scheme of Shenzhen Pengrui Foundation.

\section*{Impact Statement}
This paper presents work whose goal is to advance the field of machine learning. There are many potential societal consequences of our work, 
none of which we feel must be specifically highlighted here.

\newpage
\nocite{langley00}

\bibliography{main}
\bibliographystyle{icml2026}

\end{document}